# Corporate Evidential Decision Making in Performance Prediction Domains

A.G. Büchner*, W. Dubitzky*/‡, A. Schuster*, P. Lopes§, P.G. O'Doneghue†,
J.G. Hughes‡, D.A. Bell‡, K. Adamson‡, J.A. White¶, J.M.C.C. Anderson¥, M.D. Mulvenna*

* Northern Ireland Knowledge Engineering Laboratory, University of Ulster, U.K.
± Faculty of Informatics, University of Ulster, U.K.
§ Dept. of Physical Education, Sport and Exercise Science Unit, University of Wales, U.K.
‡ Northern Ireland Bio-Engineering Centre, University of Ulster, U.K.
¥ School of Electrical and Mechanical Engineering, University of Ulster, U.K.
¶ Dept. of Public Health Medicine and Epidemiology, Queen's Medical Centre, Nottingham, U.K.

## Abstract

Performance prediction or forecasting sporting outcomes involves a great deal of insight into the particular area one is dealing with, and a considerable amount of intuition about the factors that bear on such outcomes and performances. The mathematical Theory of Evidence offers representation formalisms which grant experts a high degree of freedom when expressing their subjective beliefs in the context of decision-making situations like performance prediction. Furthermore, this reasoning framework incorporates a powerful mechanism to systematically pool the decisions made by individual subject matter experts. The idea behind such a combination of knowledge is to improve the competence (quality) of the overall decision-making process. This paper reports on a performance prediction experiment carried out during the European Football Championship in 1996. Relying on the knowledge of four predictors, Evidence Theory was used to forecast the final scores of all 31 matches. The results of this empirical study are very encouraging.

## 1 INTRODUCTION

Predicting performance and forecasting scores in any discipline has been the subject of intensive research in the areas of traditional statistics, heuristics, or stochastic. Most common approaches use large amounts of data and apply statistical techniques. Examples are (McGarry 1994)'s model based on Markov chains used to predict squash results, (Eom 1992)'s approach which employs statistical log-linear analysis to forecast volleyball team performance, and statistical schemes such as (Hopkins 1977)'s factor analysis for basketball skill tests. All these endeavours are based on statistical schemata, and thus, have three major pitfalls in common. Firstly, human expertise or knowledge is not explicitly reflected in such prediction models, and these models usually do not account for uncertainty effects. Secondly, data is commonly collected for a pre-defined set of fixed criteria; this does not necessarily mirror the view of an individual specialist who might want to consider several hypotheses. Thirdly, the set-up criteria are always designed for a single discipline, which severely limits the versatility of such approaches. These three drawbacks also manifest themselves in many of the latest (commercial) knowledge based systems. Examples include IBM's Advanced Scout (Bhandari 1997) which analyses basketball performance and SportsOracle which predicts results of the National Football League NFL (SportsOracle 1996).

A more sophisticated performance prediction model should be able to handle knowledge originating from a number of domain experts taking into account uncertainty arising from imprecisely and partially known data. We believe that an AI approach has some advantages over statistical methods (Dubitzky 1996a/b, Chen 1992). The data in prediction models is often fraught with uncertainty which may manifest itself in the form of non-quantifiable information (subjective judgement of an individual), incomplete information (caused by inexact measurement), non-obtainable information (data that is too expensive to be established), and partial ignorance (partially known facts about a phenomenon). To make such inherently imprecise data amenable to statistical models, they must be forced into a simple numerical format, thus, at best, distorting, and at worst, destroying the reliability of the model. Furthermore, because statistics is totally data-driven, it precludes the use of available domain knowledge.

Our approach to predicting performance outcomes differs from recent work in that we incorporate expert knowledge from multiple human sources. AI offers a range of techniques to deal with uncertainty — Bayesian Networks, Fuzzy Set Theory, Rough Set Theory, Neural Networks, Genetic Algorithms, Mathematical Theory of Evidence, etc. In the performance prediction field it is often the case that many experts are readily available to provide their expertise. However, it is usually difficult to arrive at a consensus when a multitude of individual predictions are provided. This is, of course, partly caused by the subjective bias of each individual prognosticator, and by the incompleteness of the knowledge held by the predictors. The Mathematical Theory of Evidence Theory (or simply Evidence Theory) offers systematic techniques to model such scenarios. It has therefore been chosen as the decision-making framework for performance prediction.

The outline of the paper is as follows. In Section 2, we briefly recapitulate the philosophy of performance prediction and Evidence Theory respectively. In Section 3, Evidence Theory is applied to performance prediction in general. In Section 4, we set up an experiment and describe its implementation. The experiment was carried out during the European Football Championship 1996 (Euro'96)[1], and predicts 90 minute score for all matches. We then describe and evaluate the results of the experiment, before conclusions are drawn and further work is outlined.

## 2  SOME BACKGROUND

### 2.1 PERFORMANCE PREDICTION

Performance prediction is one field of sports science in which diverse disciplines are amalgamated, such as statistics, physics, and psychology. Philosophically, *performance* is defined as "the widely variable and idiosyncratic use an individual may ... make [of competence]" (Blackburn 1996). Cybernetically, *prediction* in this context is defined as "the conclusions drawn from the premise of available data using theories and models as a kind of syllogistic device", which includes "forecasting of the future as well as retrodicting the past" (Heylighen 1995). The *predictor* is a person with extensive experience in predicting results/outcomes in a particular performance prediction domain. For example, a soccer coach responsible for the national team of a country would be considered an experienced predictor.

Performance prediction is carried out by different classes of people for different reasons, and can be divided into 3 groups: competitors, coaches or advisors and externals. The predictions can be made on a microscopic or a macroscopic scale, depending on the class to which a predictor belongs and on the situation the prognosticator has to cope with. A microscopic prediction is carried out on a more spontaneous level, whereas a macroscopic forecast predicts a more complex and durable scenario. Examples of a microscopic situation is in a tennis match where the player attempts to predict the direction in which the opponent will move in order to 'wrong foot' the competitor, or in a soccer match where the goalkeeper must predict where the penalty taker is going to kick the ball (Franks 1996). Coaches and externals usually act on a macroscopic level in which patterns of play, formation and tactics of an opponent are considered as main criteria. These predictions are of vital importance in both team and individual sports, and they heavily influence the decision-making process of the forecaster. A coach or advisor employs the most appropriate pre-event training drills to synthesise the opponents' behaviour and prepare the competitor for these situations. Examples include a boxer's style, e.g., Southpaw, or the formation of a soccer team, e.g., 3 – 4 – 3. Externals, e.g., a spectator betting money on the result, or a bookmaker, setting the odds of a forthcoming event, use similar criteria to forecast the outcome of sports events.

### 2.2 MATHEMATICAL THEORY OF EVIDENCE

Evidence Theory was first put forward by Shafer and has since been extended by others (Shafer 1976, Guan 1991). The main advantages of Evidence Theory over other approaches is its ability to: (1) model the narrowing of a hypothesis set with the accumulation of evidence (via the evidence combination operation called orthogonal sum), (2) explicitly represent uncertainty in the form of ignorance or reservation of judgement, and (3) handle the reliability of the information source (experts) by means of the discount operation.

Evidence Theory describes decision problems by a set whose elements are viewed as (exhaustive and mutually exclusive) *basic hypotheses* or propositions. This set of basic hypotheses is usually symbolised by $\Theta$ and referred to as *frame of discernment*.

For instance, the frame $\Theta_{CM}$ = {*Toyota*, *Nissan*, *VW*, *BMW*} may represent alternatives for the leading car manufacturer in the US car market in 1998. The "decision" here would be to decide which is best supported by the available evidence.

In Evidence Theory belief is conceived as a quantity that can be split up, moved around and re-combined. Given a piece of evidence $e$, one can express and distribute one's belief in groups of hypotheses $X$ of $\Theta$ by $m_e(X)$, for $X \subseteq \Theta$. Because belief may be allotted to all non-empty subsets $X$ of $\Theta$, i.e., to all elements of $2^\Theta - \varnothing$, the effective hypothesis space is enlarged from $|\Theta| - 1$ to $2^{|\Theta|} - 1$.

A belief distribution over $2^{|\Theta|}$ is known as a basic probability assignment or mass function $m$; formally, $m$ is defined as follows:

$$m : 2^\Theta \to [0,1] \qquad (1)$$

such that $m(\varnothing) = 0$ and $\sum_{X \subseteq \Theta} m(X) = 1$, for all $X \subseteq \Theta$.

---

[1] The term football is homonymously world-wide used (Europeans refer to it as 'football' and Americans as 'soccer'). Throughout this text it is referred to as 'soccer'; 'football' is only used as part of event names or leagues.

When new evidence about the investigated problem emerges, it is necessary to update one's beliefs. Evidence Theory handles this accumulation or combination of evidence via the *orthogonal sum* operation (denoted by the symbol $\oplus$). For instance, the mass function $m = m_{e1} \oplus m_{e2}$ represents the combined effect of $m_{e1}$ and $m_{e2}$ based on two independent pieces of evidence $e_1$ and $e_2$. The orthogonal sum operation for $n = 2$ is given below in equation (2)

$$m(C) = \frac{\sum_{X \cap Y = C} m_{e1}(X) \cdot m_{e2}(Y)}{1 - \sum_{X \cap Y = \varnothing} m_{e1}(X) \cdot m_{e2}(Y)} \quad (2)$$

The denominator in equation (2) is called the *normalisation factor*.

The mass function can be 'discounted' by means of the so-called discount operation. Let $m_e$ be a mass function on $2^\Theta$. Given a real number $\alpha$ and a proper subset $X$ of $\Theta$, the mass function $m_e^\alpha$, defined by

$$m_e^\alpha(X) = (1-\alpha)m_e(X), \quad m_e^\alpha(\Theta) = (1-\alpha)m_e(\Theta) + \alpha \quad (3)$$

($\forall X \subset \Theta, \alpha \in [0,1]$), is said to be the discounted mass function with rate $\alpha$ of mass function $m_e$ (Guan 1991).

The discount operation is used to take into account the reliability of the information source providing the belief distributions; higher $\alpha$-values indicate lower degree of reliability.

## 3  PERFORMANCE PREDICTION AND EVIDENCE THEORY

Based on the concepts outlined in Section 2.1, the objective of performance prediction can be described as forecasting a future outcome or result as accurately as possible. Accuracy, however, can be measured on different granularity levels, depending on the type of discipline. For example, when predicting volleyball results, there exist three prediction granularity stages (winning team — set outcome — explicit result of every single set), whereas in soccer only two reasonable granularity levels exist (win, draw, loose — exact result of the match). Obviously, the number of possible outcomes increases exponentially with respect to the level of fragmentation, and thus, the more difficult its prediction becomes. In addition to the level of segmentation, the second major input factor is the quality and quantity of available expertise in the form of knowledge. It is obvious that expertise of high quality leads to more reliable performance prediction. Quantities of high quality expertise, should even lead to increasingly better forecasts, if - and this has been the dilemma so far - there is a technique of combining individual predictions.

As delineated in Section 2.2, Evidence Theory provides powerful mechanisms to handle uncertainty, which can be applied to predict performances. To be more specific, the decision problem here is that of choosing a 'best supported' basic hypothesis from the set of basic hypotheses of the form "A is better than B", "B is much better than A", and "A and B are the same" (there are a number of variations of this last hypothesis).

To demonstrate this, key features are mapped from one discipline onto the other, which is illustrated in Figure 1.

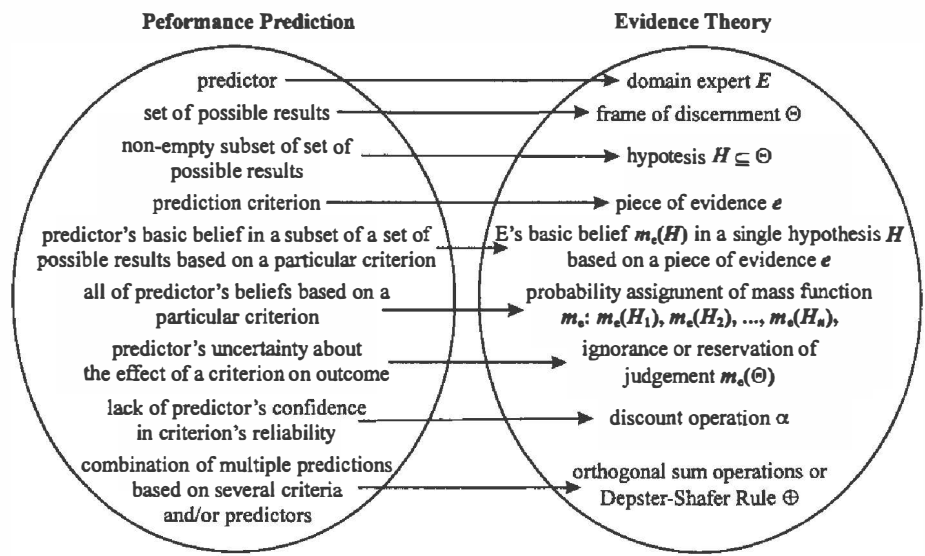

Figure 1: Mapping Performance Prediction Features onto Evidence Theory

The components of the mapping are explained in the following list:

- As defined in Section 2.1, a *predictor* can be viewed as a domain expert, e.g., an experienced soccer coach.

- The set of possible results can be seen as all *considered* possible scores of a soccer competition. The term considered refers to the closed-world assumption. This means that although a score of for example, 10:10 is in principle possible, in a competition like Euro'96 its occurrence is extremely unlikely. Such a score is therefore not considered. An example of a realistic frame of discernment is $\Theta = \{1:0, ..., 0:0, ..., 0:5\}$, and this is the one we have chosen.

- Each non-empty subset of the set of all considered possible scores is equivalent to the hypothesis notion in Evidence Theory. Examples for hypotheses include $H_1 = \{1:0, 2:1\}$, $H_2 = \{3:3, 4:4, 5:5\}$, and $H_3 = \{1:3, 1:4\}$.

- A prediction criterion is a piece of information that the predictor believes to have a significant impact or influence on the final score of a soccer match. This notion of a criterion corresponds to the piece of evidence concept in Evidence Theory. An example piece of evidence is *missing key players* (*mkp*).

- To express the degree of belief in the results in $H$ (the set of final scores or basic hypotheses) of a soccer match in conjunction with a certain criterion, a predictor associates a number from the unit interval with $H$. For example, based on the *missing key players* evidence the belief in the hypothesis $\{1:0, 2:0\}$ could be $m_{mkp}(\{1:0, 2:0\}) = 0.10$.

- A particular criterion may give rise to beliefs in several distinct score sets. Thus, the predictor has a flexible means of capturing the uncertain relationship between a specific piece of information (criterion) and the outcome of a soccer match according to his or her subjective belief. This mechanism corresponds to the basic belief or mass function in Evidence Theory, e.g., $m_{mkp}$: $m_{mkp}(\{1:0\ 2:0\}) = 0.50$ $m_{mkp}(\{2:1, 3:0, 2:1\}) = 0.20$, $m_{mkp}(\{0:0\ 1:1\}) = 0.10$, $m_{mkp}(\Theta) = 0.20$.

- A predictor may not know enough information to guage the exact influence of a certain criterion on the final score, so he or she may want reserve some of his or her judgement. This is expressed as the 'remaining' belief quantity allotted to the frame of discernment, for instance, $m_{mkp}(\Theta) = 1 - \sum_{H \subset \Theta} m_{mkp}(H) = 0.20$.

- An impact $i = 1 - \alpha$, which ranges from 0 (absolutely insignificant) to 1 (highest possible impact) can be assigned to a piece of evidence. For example, the impact of the *missing key players* evidence: $i = 0.60$. In the resulting mass function $m$ the impact $i$ is then processed as discount factor as follows $m_e^{\alpha=1-i}$.

- When all participating predictors have submitted their predictions (in the form of mass functions), the aim is to determine the most likely outcome. This is done through the application of the orthogonal sum which is defined in equation (2). See 6. Conclusions & Further Work on the issue of the independence of evidence.

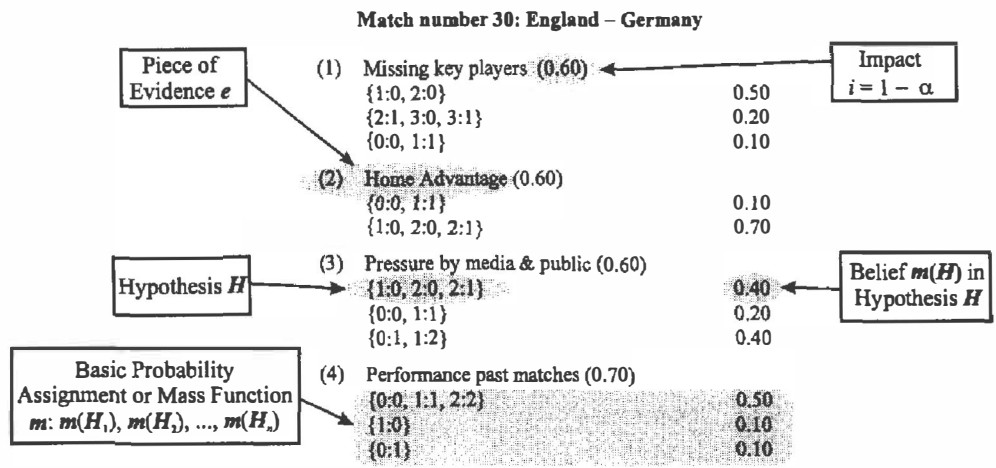

Figure 2: Sample Prediction

# 4   METHOD AND IMPLEMENTATION

First, we describe the experiment itself and the human expertise involved. Then the computation of predictions, i.e., the application of Evidence Theory in the context of performance prediction is outlined, supported by illustrative examples, and the decision-making technique which has been applied on the computed predictions is presented. Finally, the implementation of the system is outlined.

## 4.1   METHOD OF THE EXPERIMENT

A team consisting of four soccer experts has been chosen; the chosen number is completely arbitrary. The objective was to predict 90 minutes' results of all the 31 matches in Euro'96 based on the experts' knowledge. This method allowed information already gathered within the tournament, i.e., for previous matches, to be incorporated as evidence for following predictions.

For every game each expert provided his predictions (pieces of evidence) as outlined in Section 3. To illustrate the concept of the experiment an example of a prediction from one expert for match 30 is given in Figure 2.

Every prediction consists of pieces of evidence which were chosen by each expert independently. Each piece of evidence is given an impact factor $1 - \alpha$. Every piece of evidence associates a number of basic hypotheses, e.g., {2:1, 3:1, 1:0}, with a chosen belief. The set of hypotheses and its corresponding beliefs per piece of evidence build the mass function $m$. The sum of all beliefs in one basic probability assignment must not exceed one.

## 4.2   COMPUTATION OF PREDICTIONS

The frame of discernment $\Theta$ is constant throughout the whole experiment (closed-world assumption):

$$\Theta = \{ 0:0, 1:0, 2:0, 3:0, 4:0, 5:0,$$
$$0:1, 1:1, 2:1, 3:1, 4:1, 5:1,$$
$$\ldots$$
$$0:5, 1:5, 2:5, 3:5, 4:5, 5:5 \}$$

The cardinality of our frame of discernment $\Theta$ is $|\Theta| = 36$, and therefore $2^{|\Theta|} = 2^{36}$, which means that $2^{36} - 1$ possible basic hypotheses exist in the experiment. In our experiment it can be assumed that the number $n$ of chosen hypotheses $H_1, H_2, \ldots, H_n$ is always smaller than the number of possible hypotheses, i.e., $n < 2^{|\Theta|} - 1$.

For a single predictor and a particular piece of evidence $e_1$, a mass function $m_{e1}$ is described as (compare also with the evidence *missing key players* in Figure 2):

$$m_{e1} : m_{e1}(\{1:0, 2:0\}) = 0.50,$$
$$m_{e1}(\{2:1, 3:0, 3:1\}) = 0.20,$$
$$m_{e1}(\{0:0, 1:1\}) = 0.10,$$
$$m_{e1}(\Theta) = 0.20$$

Applying the orthogonal sum to $m_{e1}$, $m_{e2}$, $m_{e3}$, and $m_{e4}$, according to equation (2), yields the combined belief distribution $m_c$ (which can be seen in Table 1):

$$m_c = m_{e_1} \oplus m_{e_2} \oplus m_{e_3} \oplus m_{e_4}$$

Table 1: Combination of Evidence

| # | Hypothesis | Belief |
|---|---|---|
| 1 | {1:0, 2:0} | 0.377 |
| 2 | {1:0} | 0.176 |
| 3 | {2:1} | 0.151 |
| 4 | {1:0, 2:0, 2:1} | 0.151 |
| 5 | {0:0, 1:1} | 0.101 |
| 6 | {0:1, 1:2} | 0.034 |
| 7 | {0:1} | 0.001 |
| 8 | {2:1, 3:0, 3:1} | <0.001 |
| 9 | {0:0, 1:1, 2:2} | <0.001 |

This table can be interpreted as follows: Out of $2^{36} - 1$ possible hypotheses, 9 have been chosen by the expert; the belief in each of them is given in the rightmost column in order of their importance.

The objective of the application of Evidence Theory is the prediction of one single result (basic hypothesis), and not a set of results (group of basic hypotheses). This issue is addressed in the following section.

## 4.3   DECISION-MAKING

It is possible that an outcome contains a set of results with more than one element (see Table 1). Although it may seem obvious from the first two hypotheses in this example that 1:0 is the result with the highest belief, this method does not have a theoretical foundation. To conform to the underlying philosophy of Evidence Theory, i.e., all decisions must be grounded on evidence and no other information, all experts have to *update* their beliefs based on new evidence to come to an unambiguous conclusion. This method may seem obvious for some scenarios, e.g., medical applications, but can be awkward in other situations, e.g. time-critical decision-making. For the purpose of our experiment, we decided that a fully automated solution is more practical, and calculated the distribution of belief over singletons using traditional statistical techniques, applying the weighted distribution of all sets to atomic results. The result of the distribution of belief over singletons is as follows[2]:

---

[2] More sophisticated techniques to solving this problem exist, such as (Xu 1995)'s sensitivity analysis.

Table 2: Distribution of Belief over Singletons

| # | Result | Belief | # | Result | Belief |
|---|--------|--------|---|--------|--------|
| 1 | 1:0 | 0.415 | 6 | 0:1 | 0.028 |
| 2 | 2:0 | 0.239 | 7 | 1:2 | 0.017 |
| 3 | 2:1 | 0.201 | 8 | 3:0 | <0.001 |
| 4 | 0:0 | 0.050 | 9 | 3:1 | <0.001 |
| 5 | 1:1 | 0.050 | 10 | 2:2 | <0.001 |

After the application of the distribution of belief over singletons, we now have a ranking of atomic results, which shows that the score 1:0 has the highest belief of 41.5%. We would like to stress that in our experiment a distribution was rarely necessary, but when applying Evidence Theory to other sporting disciplines with a smaller frame of discernment, the application might be more often required.

Table 3: Individual Evidence Theory based predictions $d_e$

| Expert | Correct Results | Correct Outcomes | $\sum s_1$ | $\frac{1}{n}\sum s_1$ | $\sum s_2$ | $\frac{1}{n}\sum s_2$ | $\sum s_3$ | $\frac{1}{m}\sum s_3$ |
|--------|-----------------|------------------|------------|-----------------------|------------|-----------------------|------------|-----------------------|
| $e_1$ | 4 | 12 | 4 | **0.129** | 16 | 0.526 | 20 | 0.323 |
| $e_2$ | 4 | 7 | 4 | **0.129** | 11 | 0.355 | 15 | 0.242 |
| $e_3$ | 3 | 10 | 3 | 0.097 | 13 | 0.429 | 16 | 0.258 |
| $e_4$ | 2 | 10 | 2 | 0.097 | 12 | 0.387 | 14 | 0.226 |
| $\frac{1}{4}\sum_{i=1}^{4} e_i$ | 3.25 | 9.75 | 3.25 | 0.113 | 13.00 | 0.422 | 16.25 | 0.262 |

Table 4: Individual intuition based predictions $d_i$

| Expert | Correct Results | Correct Outcomes | $\sum s_1$ | $\frac{1}{n}\sum s_1$ | $\sum s_2$ | $\frac{1}{n}\sum s_2$ | $\sum s_3$ | $\frac{1}{m}\sum s_3$ |
|--------|-----------------|------------------|------------|-----------------------|------------|-----------------------|------------|-----------------------|
| $e_1$ | 5 | 10 | 5 | **0.161** | 15 | **0.484** | 20 | 0.323 |
| $e_2$ | 2 | 13 | 2 | 0.065 | 15 | **0.484** | 17 | 0.274 |
| $e_3$ | 3 | 10 | 3 | 0.097 | 13 | 0.419 | 16 | 0.258 |
| $e_4$ | 2 | 12 | 2 | 0.065 | 14 | 0.452 | 16 | 0.258 |
| $\frac{1}{4}\sum_{i=1}^{4} e_i$ | 3.00 | 11.25 | 3.00 | 0.097 | 14.25 | 0.460 | 17.25 | 0.278 |

### 4.4 IMPLEMENTATION

A complete application has been implemented to compute results based on Evidence Theory for Euro'96 (Figure 3). Evidence from four experts was collected, stored electronically, and used as input for an evidential reasoning algorithm, which has been developed in C++. Dependent on the result sets and distribution of belief over singletons, a decision was made.

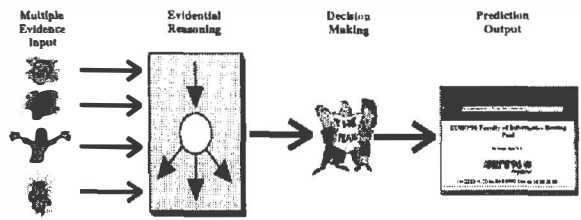

Figure 3: Performance Prediction System

## 5 RESULTS AND EVALUATION

First, assessment criteria for the experiment are described, before the actual results are presented. Then the results are evaluated corresponding to assessment criteria set up in the following sub-section.

### 5.1 ASSESSMENT CRITERIA

The main assessment criterion was not a particular percentage of correct predictions because the choice of experts was of a purely subjective nature. The most valuable evaluation is how the quality of experts' intuitive predictions relate to experts' predictions based on Evidence Theory. Hence, in addition to providing pieces of evidence, all experts were asked to submit their intuitive prediction, which leads to the following available data for each match and expert, respectively:

$d_i$  Intuitive prediction of each individual expert
$d_e$  Evidence Theory based prediction of each individual expert
$d_c$  Combined predictions of all experts using Evidence Theory — as an interesting experiment

To be compatible, for interest, with as many betting systems as possible, the following performance prediction schemata were used for evaluation:

$s_r$  Results only, i.e., the exact result of a match. This prediction is used in many internationally operated betting systems.

- $s_o$  Outcomes only, i.e., a hit in the set {win, draw, loose}. This is how the Toto systems, played in several countries work[3].

- $s_{ro}$  Results and outcomes, i.e., 2 points for the right result and 1 point for the right outcome. This schema is a mixture of $s_r$ and $s_o$.

The final criterion is the qualitative spectrum of the expertise team, which could lead to manipulated results. For instance, one top expert with a very high number of correct results, could smooth out participating weaker experts, or vice versa. Thus, the two statistical functions of *max* and *mean* have been applied to the data and schema mentioned above.

## 5.2 RESULTS

Table 3 and Table 4 show the results of the intuition based predictions $d_i$ and Evidence Theory based predictions $d_e$. For each expert $e_1, ..., e_4$ correct results and outcomes are given in columns 2 and 3, respectively. Columns 4, 6 and 8 represent the schemata $s_r$, $s_o$, and $s_{ro}$. Columns 5, 7 and 9 show the ratio between scored points and possible points (*n* denotes the number of 31 matches, and m the number of possible 62 points). The *mean* over the single columns has been calculated in the last row, *max* values are printed in bold.

The combined Evidence Theory based results $d_c$ are summarised in Table 5, where formulae denote identical headings as in the previous two tables[4].

Table 5: Combined Evidence Theory based Predictions $d_c$

| $s_r$ | $s_o$ | $\sum s_1$ | $\frac{1}{n}\sum s_1$ | $\sum s_2$ | $\frac{1}{n}\sum s_2$ | $\sum s_3$ | $\frac{1}{m}\sum s_3$ |
|---|---|---|---|---|---|---|---|
| 6 | 10 | 6 | 0.194 | 16 | 0.526 | 22 | 0.355 |

## 5.3 EVALUATION OF RESULTS

The results shown in tables 3 to 5 can be summarised as follows. 'Correct Results' predictions based on combined evidence (19.4%) are better than the average individual intuition (9.7%) and evidence based (11.3%) forecasts, and also better than the best expert's forecast (16.1% and 12.9%). 'Correct Outcome' predictions show a very similar behaviour, with one subtle difference, which is the identical prediction of the best expert based on evidence theory (52.6%).

Using prediction schema $s_{ro}$, which is based on outcomes *and* results, a similar behaviour to schema $s_r$ can be stated, although the differences between combined evidence theory based and individual predictions are slightly smaller. The summarised interpretation is depicted in Figure 4.

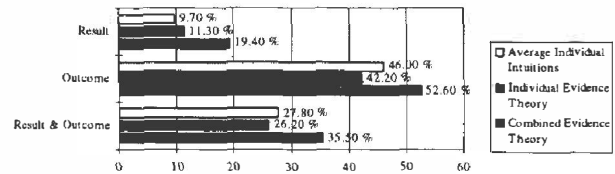

Figure 4: Individual / Evidence Theory based Predictions

The provided hypotheses consist of results, rather than outcomes, although the outcome of a match is implicit, e.g., the hypothesis {0:0, 1:1, 2:2} explicitly states a draw as outcome, but does not precisely specify which draw. This can be accepted as an explanation for better result predictions, than for pure or mixed outcome predictions ($s_o$ and $s_{ro}$). It is believed that considering outcomes explicitly in the frame of discernment, i.e., $\Theta_{outcome} = \{win, draw, loose\}$, would have led to similar encouraging results for schemata $s_o$ and $s_{ro}$.

## 6  CONCLUSIONS & FURTHER WORK

We have presented a study of the application of Evidence Theory for performance. Setting up a real-world experiment, namely the prediction of results of all Euro'96 matches, the theory provided good predictions. The results are solid demonstrations of the advantages of Evidence-Theory-based performance prediction over traditional statistical models — diverse expertise can be considered, unknown information can be handled, and knowledge can be combined. Although the presented performance prediction method has been applied to soccer results, it seems to be applicable to other sports disciplines.

Combining hypotheses from different experts using the orthogonal sum operation ⊕ has shown a valuable improvement in the quality of performance prediction over the individual predictions for this admittedly anecdotal study. To show the scalability of Evidence Theory, individual hypotheses of different numbers of specialists have to be considered and evaluated in the same experiment using the methods and assessment criteria outlined above. Clearly, it cannot be expected that the prediction quality increases linearly with respect to the number of experts, and thus with the number of mass functions, but it can be expected that it will increase to some extend. How much has to be explored in future research.

A common critique of reasoning under uncertainty using Evidence Theory is its limitation that pieces of evidence must be independent (Ling 1989b). For example, two pieces of evidence $e_1$ = 'home advantage' and $e_2$ = 'expected strategy' influence each other to a certain degree, but are treated as if they are be independent of each other, and thus have more impact on the prediction than they should have. (Ling 1989a) tackled the problem on an epistemological level and introduced parameters which quantify the degree of dependence between and among pieces of evidence. The problem in this approach is that the weight of each dependency has to be estimated (who decides, and how, on the dependence of $e_1$ and $e_2$?). Another attempt by (Ling 1989b) involves the combination of knowledge at a statistical level. This approach is not applicable to the outlined performance prediction scenario, because it assumes that "precise mathematical information about covariances between bodies of evidence is available".

---

[3] This system has different names in other countries, such as Pools or Tote

[4] The results can also be viewed in the statistics section under the URL http://www.infj.ulst.ac.uk/~cbgv24/euro96/index.html.

In the evaluation of the experiment in Section 5, the correct results and outcomes of the matches were used to evaluate the method. This does, strictly speaking, contradict our experimental set-up in Section 4. But, viewing the scenario from an abstract point of view, predicting soccer results can be sub-divided into two mindsets: an outcome mindset and a result mindset. The result mindset, in which explicit results are predicted can be coarsened to the outcome based mindset, in which only outcomes are predicted. Of course, more levels could be imagined in this hierarchy, such as {*clear win, normal win, close win*}. The procedure, described in detail by (Guan 1992), is bi-directional, i.e., refining outcome based predictions to result based predictions is also possible. Introducing explicit coarsening and refinement operations promises to be a valuable extension for more accurate performance prediction.